\documentclass[runningheads]{llncs}
\usepackage{graphicx}
\usepackage{amsmath,amssymb} % define this before the line numbering.
\usepackage{color}
\usepackage{bm}
\usepackage{amsfonts}
\usepackage{subfig}
\usepackage{multirow}
\usepackage{makecell}
\usepackage{wrapfig,lipsum,booktabs}
\newcommand\norm[1]{\left\lVert#1\right\rVert}

\begin{document}
%===========================================================

\title{Traversing Latent Space using Decision Ferns\thanks{This work was supported by the Australian Research Council Centre of Excellence for Robotic Vision (project number CE1401000016).}} % Replace your paper's title here
\titlerunning{Traversing Latent Space using Decision Ferns} % Replace an abstracted version of your paper's title here

%===========================================================

\author{Yan Zuo\thanks{Authors contributed equally} \and
Gil Avraham$^{\star\star}$ \and
Tom Drummond}
%
%Please include author names in full in the paper, 
%If any authors have names that can be parsed into FirstName LastName in multiple ways, please include the correct parsing, in a comment to the volume editors:
%\index{Lastnames, Firstnames}

\authorrunning{Zuo et al.} % A shorter version of authors' name
% First names are abbreviated in the running head.
% If there are more than two authors, 'et al.' is used.

%===========================================================

\institute{ARC Centre of Excellence for
Robotic Vision, Monash University, Australia \\
\email{yan.zuo@monash.edu}, \email{gil.avraham@monash.edu}, \email{tom.drummond@monash.edu}}

\maketitle

%===========================================================
\begin{abstract}
The practice of transforming raw data to a feature space so that inference can be performed in that space has been popular for many years. Recently, rapid progress in deep neural networks has given both researchers and practitioners enhanced methods that increase the richness of feature representations, be it from images, text or speech. In this work we show how a constructed latent space can be explored in a controlled manner and argue that this complements well founded inference methods. For constructing the latent space a Variational Autoencoder is used. We present a novel controller module that allows for smooth traversal in the latent space and construct an end-to-end trainable framework. We explore the applicability of our method for performing spatial transformations as well as kinematics for predicting future latent vectors of a video sequence.
\end{abstract}

%===========================================================
\section{Introduction}
\label{sec:intro}
A large part of human perception and understanding relies on using visual information from the surrounding environment and interpreting this information to subsequently act upon these interpretations. However, these interpretations are not applied directly to what is being seen; rather, there exists an abstraction mechanism from the image space into a more informative space so that complex inferences can be made there~\cite{utgoff2002many}. Similarly in machine learning, we would like machines to inherit this ability to abstract as it is the key to understanding and learning when it comes to real world data.

Recently, deep learning has demonstrated enormous success across various vision-related tasks such as image classification, object detection and semantic segmentation~\cite{krizhevsky2012imagenet,simonyan2014very,he2016deep,girshick2014rich,long2015fully}. However, this level of success has yet to transition across to more complicated tasks such as video prediction. Many of the popular deep learning methods approach these more challenging tasks in a similar manner to image classification or segmentation, choosing to learn directly from the image space~\cite{lotter2016deep,mathieu2015deep}. This presents a challenge because often the image space is high-dimensional and complex; there exists a large semantic gap between the input pixel representation of the data and the desired transformation of said data for complex inference tasks such as prediction.

To address this challenge, we leverage the compact encoding space provided by a Variational Autoencoder (VAE)~\cite{kingma2013auto} to learn complex, higher order tasks in a more feasible way. Inference in the latent space takes advantage of solving a far more tractable problem than performing the operation in the image space as it gives a more natural way of separating the task of image construction and inferring semantic information. Other works have similarly utilised a compact encoding space to learn complex functions~\cite{liang2017dual,yoo2017variational,kulkarni2015deep,santana2016learning}; however, even this encoding space can be strongly entangled and highly non-linear. As such, we construct a residual decision fern based architecture which serves as a controller module that provides the necessary non-linearity to properly disentangle encodings in the latent space. To this end, we introduce a novel framework for controlled traversal of the latent space called the Latent Space Traversal Network (LSTNet) and offer the following contributions:

\begin{itemize}
\item We discuss the benefits of operating in a latent space for complex inference tasks, introducing a novel decision fern based controller module which enables the use of control variables to traverse the latent space (Sections~\ref{ssec:constructing_latent_space},~\ref{ssec:traversing_latent_space} \&~\ref{ssec:lstnet}).
\item We create a unified, end-to-end trainable framework which incorporates our controller module with a residual VAE framework, offering an encoding space for learning high order inference tasks on real world data. Additionally, this framework offers a key insight into separating the tasks of pixel reconstruction and the high order inference (Sections~\ref{ssec:lstnet} \&~\ref{ssec:latent_space_for_prediction}).
\item We demonstrate significant qualitative and quantitative improvements offered by LSTNet over popular models that impose geometrical and kinematic constraints on the prediction search space across the MNIST and KITTI datasets (Sections~\ref{ssec:imposing_spatial},~\ref{ssec:imposing_kinematics} \&~\ref{ssec:latent_space_for_prediction}). 
\end{itemize}

%-----------------------------------------------------------------
\section{Related Work}
\subsection{Learning Representations for Complex Inference Tasks}
\label{ssec:vaerelated}
Operating in the latent space for demonstrating certain properties has been shown in several works. Applying a convolution architecture to a GAN framework~\cite{radford2015unsupervised} showed that following the construction of the latent space and applying arithmetic operations between two latent vectors observes the semantic logic we would expect from such an operation. InfoGAN~\cite{chen2016infogan} added a regularisation term which maximises mutual information between image space and some latent variables. Both these works claim to yield a disentangled latent vector representation of the high dimensional input data, demonstrating this by choosing a specific latent variable and interpolating across two values and showing smooth image transformation. However, due to their unsupervised nature, there are no guarantees on what attributes they will learn and how this will distort the intrinsic properties of the underlying data. 

The work of ~\cite{kulkarni2015deep} divides the learned latent space in a VAE into extrinsic and intrinsic variables. The extrinsic variables are forced to represent controllable parameters of the image and the intrinsic parameters represent the appearance that is invariant to the extrinsic values. Although it is trained in a VAE setting, this method requires full supervision for preparing training batches.~\cite{yoo2017variational} introduces a fully supervised method, introducing a Gaussian Process Regression (GPR) in the constructed latent space of a VAE. However, using GPR imposes other limitations and assumptions that do not necessarily apply in training a VAE.~\cite{santana2016learning} uses a semi-supervised approach for video prediction, training a VAEGAN~\cite{larsen2015autoencoding} instead of a standard VAE. The VAEGAN leads to sharper looking images but imposes a discriminator loss which complicates the training procedure drastically. This work is similar to ours in that control variables are used to guide learning in the latent space. However, this method is not end-to-end trainable and uses a simple framework for processing the latent space;~\cite{lotter2016deep} showed this framework underperformed in next frame prediction and next frame steering prediction.

~\cite{mathieu2015deep,liang2017dual} offered a video prediction framework which combines an adversarial loss along with an image gradient difference/optical flow loss function to perform next frame prediction and also multi-step frame prediction. Similarly,~\cite{lotter2016deep} performed predictive coding by implementing a convolutional LSTM network that is able to predict the next frame and also multi-step frame prediction in a video sequence. In contrast to our work and~\cite{santana2016learning}, these works optimise learning in the image space. This approach suffers from poor semantic inference (especially over large time intervals) and will be a focus of investigation in our work. 

\subsection{Decision Forests \& Ferns}
\label{ssec:decisionforestsrelated}
Decision forests are known for their flexibility and robustness when dealing with data-driven learning problems~\cite{caruana2006empirical}. Random decision forests were first developed by~\cite{ho1998random}, where it was found that randomly choosing the subspace of features for training had a regularising effect which overcame variance and stability issues in binary decision trees. Following this, various methods which added more randomness quickly followed which helped further stabilise training of decision trees~\cite{breiman1996bagging,breiman2001random,geurts2006extremely}. Further work extended decision trees into a related method of decision ferns, finding use in applications such as keypoint detection~\cite{ozuysal2010fast,kursa2012rferns}. Recently, an emerging trend has seen the incorporation of decision forests within deep learning frameworks~\cite{bulo2014neural,kontschieder2015deep,zuo2017fast,zuo2018generative}, utilising the non-linear discriminating capabilities offered by deep decision trees.

%-----------------------------------------------------------------
\section{Background}
\label{sec:background}
\subsection{Variational Autoencoders}
\label{ssec:vaebackground}
There are many works that approximate probability distributions. For a comprehensive review, refer to~\cite{bengio2013representation} and more recently~\cite{rosca2017variational}.~\cite{kingma2013auto} proposed a unified encoder-decoder framework for minimising the variational bound:
\begin{equation}
\begin{split}
\log{P(X)} - \mathcal{D}[\mathcal{Q}(z|X)||P(z|X)] = \\ E_{z\sim\mathcal{Q}}[log{P(X|z)}]-\mathcal{D}[\mathcal{Q}(z|X)||P(z)]\label{eq:1}
\end{split}
\end{equation} 
It states that minimising the term $\mathcal{D}[\mathcal{Q}(z|X)||P(z)]$ by choosing $z\sim\mathcal{N}(\mu,\sigma)$ and imposing a normal distribution on the output of the encoder will result in approximating $P(X)$ up to an error that depends on the encoder-decoder reconstruction error. In practice, the encoder and decoder are constructed to have sufficient capacity to encode and decode an image from the given distribution properly, and also the choice of latent space to be in a high enough dimensionality to have the descriptive capacity to encode the image information needed for reconstruction.
\subsection{Decision Forests \& Ferns}
\label{ssec:decisionfernsbackground}
% In this work, we utilise the soft decision forest proposed in~\cite{kontschieder2015deep}.
\paragraph{Decision Trees}
A decision tree is composed of a set of internal decision nodes and leaf nodes $\ell$. The decision nodes, $\mathcal{D}=\{d_0, \cdots, d_{N-1}\}$, each contain a decision function $d(\bm{x}; \theta)$, which perform a routing of a corresponding input based on its decision parameters $\theta$. The set of input samples $\bm{x}$ start at the root node of a decision tree and are mapped by the decision nodes to one of the terminating leaf nodes $\ell = \mathcal{D}(\bm{x},\Theta)$, which holds a real value $\bm{q}$ formed from the training data ($\Theta$ denotes the collected decision parameters of the decision tree). Hence, a real value $q$ in a leaf node will be denote by:
\begin{equation}
q(\ell) = \frac{\sum_i \omega(\mathcal{D}(x_i)) v_{i}}{n_\ell}
\label{eq:leaf_route_function}
\end{equation}
where $n_\ell$ is the total count of training samples mapped to leaf node $\ell$, $\omega$ specifies the decision tree mapping of the sample to leaf $\ell$ and $v_{i}$ specifies the value of sample $i$.
\paragraph{Decision Ferns}
A decision fern is related to a decision forest. Similarly to decision trees, decision ferns route samples from the root node to a leaf node $\ell$. The key difference is that all decisions are made simultaneously rather than in sequence. Thus, a decision fern contains a single root node which holds the parameters $\Theta$ of the entire decision fern to map from the root to a leaf node.

%----------------------------------------------------------------
\section{Operating in Latent Space}
\label{sec:operating_in_latent_space}
To operate in latent space, there must exist a mechanism which allows for transition between the latent space and image space. Hence, there is an inherent trade off between obtaining a good semantic representation of the image space via its corresponding latent space, and the reconstruction error introduced when transitioning back to the image space. In this work, we show that the benefits of working in a compact latent space far outweigh the loss introduced by image reconstruction. The latent space emphasises learning the underlying changes in semantics related to an inference task which is paramount when learning high order inference tasks.

\subsection{Constructing a Latent Space}
\label{ssec:constructing_latent_space}
When constructing a latent space, several viable options exist as candidates. The most naive method would be to perform Principal Component Analysis (PCA) directly on images~\cite{turk1991face}, selecting the $N$ most dominant components. Clearly, this method results in a large loss in information and is less than ideal. Generative Adversarial Networks (GANs)~\cite{radford2015unsupervised,goodfellow2014generative,arjovsky2017wasserstein} can produce realistic looking images from a latent vector but lack an encoder for performing inference and are difficult to train. Techniques which combine GANs with Variational Autoencoders~\cite{larsen2015autoencoding,rosca2017variational,dumoulin2016adversarially,donahue2016adversarial,mescheder2017adversarial} offer an inference framework for encoding images but prove to be cumbersome to train due to many moving parts and instability accompanying their adversarial training schemes.

Hence, to construct our latent space, we use the relatively straightforward framework of a Variational Autoencoder (VAE)~\cite{kingma2013auto}. VAEs contain both a decoder as well as an encoder; the latter of which is used to infer a latent vector given an image. This encoding space offers a low dimensional representation of the input and has many appealing attributes: it is semi-smooth and encapsulates the intrinsic properties of image data. It is separable by construction; it maintains an encoding space that embeds similar objects in the image space near each other in the latent space. Furthermore, a VAE can be trained in a stable manner and in an unsupervised way, making it an ideal candidate to learn complex higher order inference tasks.

\subsection{Traversing the Latent Space}
\label{ssec:traversing_latent_space}
Recent works attempted to learn the latent space transformation using various models, such as RNNs~\cite{santana2016learning} and a Gaussian Process Regression~\cite{yoo2017variational}. In our work we recognise that although the original input data is reduced to a lower dimension encoding space, inference on this space is a complex operation over a space which if constructed correctly, has no redundant dimensions; hence all latent variables should be utilised for the inference task. Under the assumption of a smooth constructed manifold and a transformation that traverses this smooth manifold under a narrow constraint (in the form of a control variable or side information), a reasonable model for the controller module is:
\begin{equation}
z_{t+h} = z_{t} + F(z_{t},z_{t-1},z_{t-2},...,\theta)
\label{eq:controller_module}
\end{equation}

where $\{z_{t},z_{t-1},z_{t-2}...\}$ are the latent vectors corresponding to input data $\{x_{t},x_{t-1},x_{t-2}...\}$, $\theta$ is the control variable and $z_{t+h}$ is the output of the model corresponding to given the inputs. The operator $F(\mathbf{z},\theta)$ can be interpreted as:
\begin{equation}
\begin{split}
\frac{z_{t+h} - z_{t}}{h} = \frac{1}{h}F(z_{t},z_{t-1},z_{t-2},...,\theta) \\
\frac{\partial{z_{t}}}{\partial{h}} = F(z_{t},z_{t-1},z_{t-2},...,\theta)\label{eq:transformer_network}
\end{split}
\end{equation}
where $\frac{1}{h}$ can be absorbed into $F(\mathbf{z},\theta)$ and by doing so, we can interpret it as a residual term that is added for smoothly traversing from input $z_{t}$ to $z_{t+h}$, given side information $\theta$ and the history $\{z_{t},z_{t-1},z_{t-2}...\}$. This construction allows us to implement Eq.~\ref{eq:transformer_network} using a neural network, which we denote as the \textit{Transformer Network}. Eq.~\ref{eq:controller_module} encapsulates the complete controller, which is a residual framework that delivers the final transformed latent vector (denoted as the \textit{Controller Module}). The Transformer Network, will be doing the heavy lifting of inferring the correct step to take for obtaining the desired result $\hat{z}_{t+h}$ and as such should be carefully modelled. In the following section, we discuss our chosen implementation and the considerations that were taken when constructing the Transformer Network.

\subsection{Latent Space Traversal Network}
\label{ssec:lstnet}
\begin{figure*}
\centering
\includegraphics[width=0.8\textwidth]{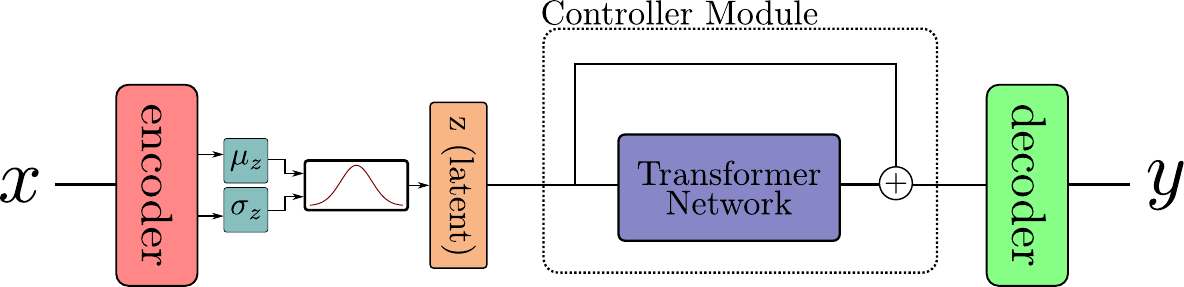}
\caption{Overview of our LSTNet model architecture. $x$ denotes input data and $y$ denotes the output with a transformation applied in the latent space.}
\label{fig:lstnet_model_arch}
\end{figure*}

Our Latent Space Traversal Network (LSTNet) consists of two main components:
\begin{enumerate}
\item A VAE with an encoder and decoder. The encoder learns a latent representation to encode the real set of training images into this space. The decoder learns a mapping from latent space back to the image space.
\item A Controller Module (C) with a Transformer Network (TN), that applies an operation in the latent space offered by the VAE.
\end{enumerate}
An overview of our model is shown in Fig.~\ref{fig:lstnet_model_arch}. In the rest of this section, we detail and justify the choice of architecture for the components of our model.
To construct our latent space, we adopt the approach in~\cite{kingma2016improved} and construct a residual encoder and decoder to form our VAE. This residual VAE offers a low-dimensional, dense representation of input image data, allowing for a latent space which makes higher order inference tasks easier to learn in. LSTNet is trained on a loss function composed of three terms:
\begin{equation}
\begin{split}
\mathbb{L}_{VAE} = \frac{1}{N} \sum_{i \in \mathcal{B}} \norm{\hat{I_{i}}-I_{i}}^2 + \frac{1}{N} \sum_{i \in \mathcal{B}} KL(z_{i}, \mathcal{N}(0,U)), \\ 
z\sim \mathcal{N}(\mu_{enc}(I),\sigma_{enc}^2(I))
\end{split}
\label{eq:vae_loss}
\end{equation}
\begin{equation}
\begin{split}
\mathbb{L}_{z} = \frac{1}{N} \sum_{i \in \mathcal{B}} \norm{z_{target}-\hat{z}_{target}}^2, \hat{z}_{target}=C(z_{t-n},...,z_{t-1},\theta), \\ 
z_{target} = \mu_{enc}(I_{t}) 
\end{split}
\label{eq:controller_loss}
\end{equation}
\begin{equation}
\begin{split}
\mathbb{L}_{I} = \frac{1}{N} \sum_{i \in \mathcal{B}} \norm{I_{target}-\hat{I}_{target}}^2, \hat{I}_{target}=\mathcal{P}(I_{t-n},...,I_{t-1},\theta), \\
I_{target} = I_{t} 
\end{split}
\label{eq:prediction_loss}
\end{equation}
where $~\mathbb{L}_{VAE}$ is the loss for the VAE which updates the encoder and decoder parameters, $~\mathbb{L}_{z}$ is the controller loss which updates the controller network's parameters and $~\mathbb{L}_{I}$ is the predicted image loss which updates the controller and decoder of the VAE. In this case, $I$ is an image, $z$ is a latent vector in the encoding space, $\mathcal{B}$ is a minibatch, $U$ is an identity matrix, $\mu_{enc}$ and $\sigma_{enc}^2$ are the respective mean and variance of the encoder's output, $C$ is the controller network and $\mathcal{P}$ denotes the LSTNet (passing an input image(s) through the encoder, controller and decoder to generate a transformation/prediction).

\paragraph{Fern-based Transformer Network}
% \label{ssec:fern_based_transformer_network}
Even in the latent space, learning complex tasks such as video prediction can be difficult. 
In the experiments section, we motivate the use of the fern-based controller over a linear variant composed of stacked fully connected layers with ReLU non-linearities.

\begin{figure*}
\centering
 \subfloat[]{%
  \includegraphics[height=0.25\textwidth]{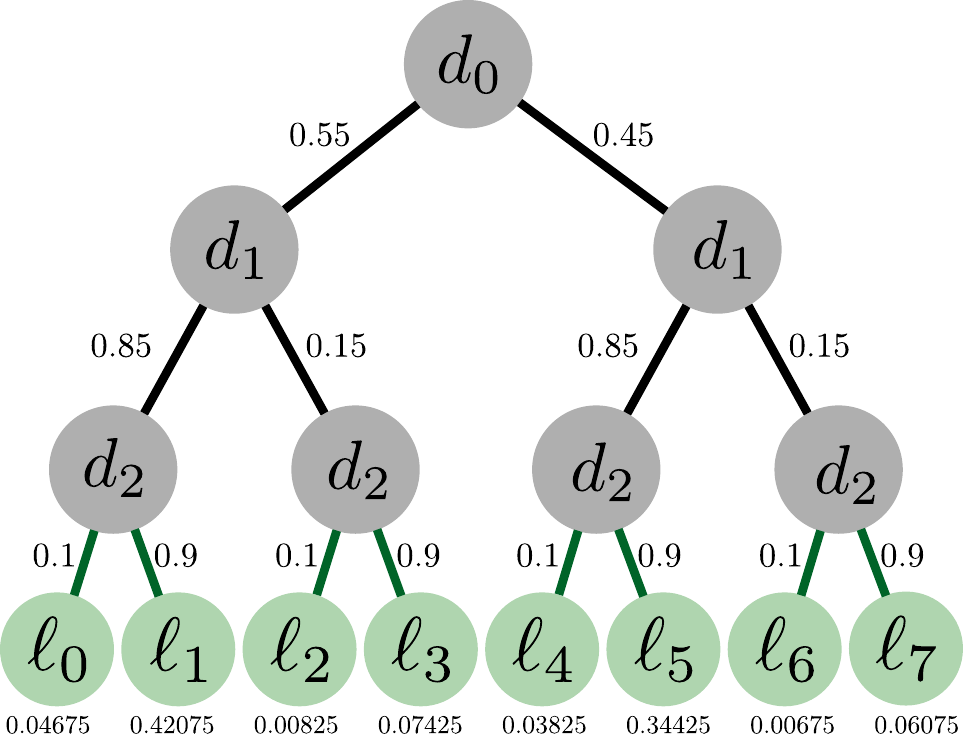}
  \label{fig:soft_fern}
 }%
 \subfloat[]{%
  \includegraphics[height=0.3\textwidth]{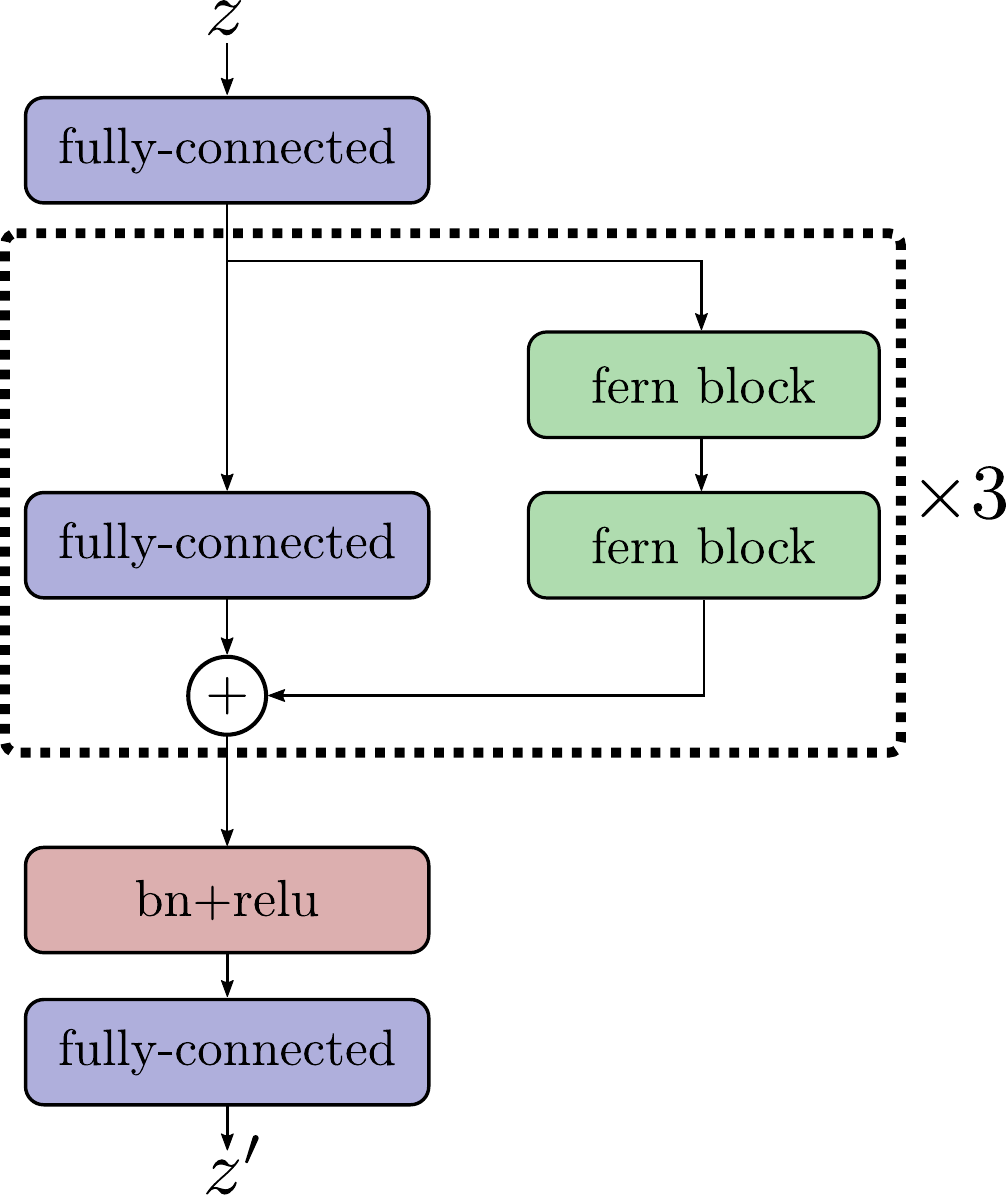}
  \label{fig:controller_arch}
 }%
 \subfloat[]{%
  \includegraphics[height=0.3\textwidth]{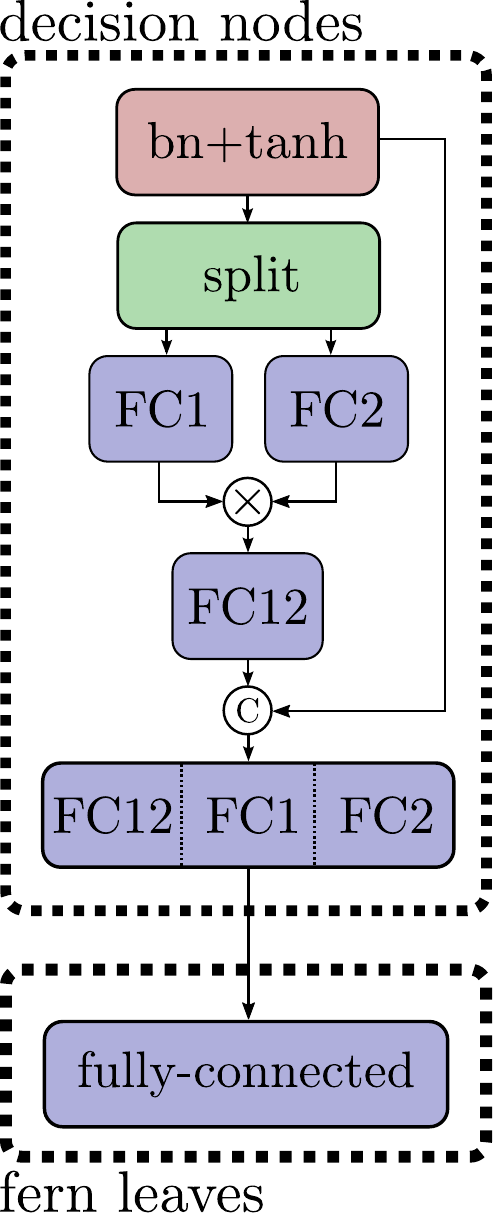}
  \label{fig:fern_block}
 }%
\caption{(a) A decision fern with soft routing. (b) The architecture of our proposed transformer network. (c) The structure of a fern block used in our transformer network.}
\label{fig:model_components}
\end{figure*}

Our transformer network employs an ensemble of soft decision ferns as a core component. The use of soft decision ferns allows them to be differentiable such that they can be integrated and trained within an end-to-end framework. One way to achieve this is construct decision functions which apply a sigmoid to each input activation biased with a threshold value, yielding a soft value between $[0, 1]$:
\begin{equation} 
d_n(\bm{x},\bm{t}) = \sigma((x_n-t_n))
\label{eq:soft_decision_function}
\end{equation}
where $\sigma(x)$ is a sigmoid function. $x_n$ and $t_n$ are the respective input activation and corresponding threshold values assigned towards the decision. To illustrate this, for a depth two fern using two activations which create the soft routes to its corresponding four leaves, its output $Q$ is:
\begin{equation}
\begin{split}
Q = q_0 \times p_0 \times p_1 + q_1 \times p_0 \times (1-p_1) \\ + q_2 \times (1-p_0) \times p_1 + q_3 \times (1-p_0) \times (1-p_1)
\end{split}
\label{eq:soft_fern_output}
\end{equation}
where $p_0$ and $p_1$ are the respective probability outputs of the decision functions of the decision ferns. $q_0, q_1, q_2$ and $q_3$ are the corresponding leaf nodes of the decision fern (illustrated in Fig.~\ref{fig:soft_fern}). Eq.~\ref{eq:soft_fern_output} can be reparameterised as:
\begin{equation}
Q = b + d_0 \times x + d_1 \times y + d_0 \times d_1 \times z
\end{equation}
where:
\begin{equation}
\begin{split}
d_0=\tanh(x_0), d_1=\tanh(x_1) \\ 
b = \frac{1}{2^h}\times(q_0+q_1+q_2+q_3), x = \frac{1}{2^h}\times(q_0-q_1+q_2-q_3) \\
y = \frac{1}{2^h}\times(q_0+q_1-q_2-q_3), z = \frac{1}{2^h}\times(q_0-q_1-q_2+q_3)
\end{split}
\end{equation}
$x_0$ and $x_1$ are the assigned activations to the decision fern and $h$ is the fern depth. $b, x, y, z$ can be represented by fully connected linear layers which encapsulates all decision ferns in the ensemble. 

Fig.~\ref{fig:controller_arch} shows the architecture of our transformer network. We adopt the residual framework in~\cite{he2016deep}, modifying it for a feedforward network (FNN) and adding decision fern blocks along the residual branch in the architecture. In Fig.\ref{fig:fern_block}, we outline the construction of a decision fern building block in the TN, consisting of ferns of two levels in depth. The decision nodes of the fern are reshaped from incoming activations, to which Batch Normalisation~\cite{ioffe2015batch} and a Hyperbolic Tangent function is applied. This compresses the activations between the range of $[-1,1]$ and changes their role to that of making decisions on routing to the leaf nodes. A split and multiply creates the conditioned depth two decisions of the fern, which is concatenated with the depth one decisions. Finally, a FC linear layer serves to interpret the decisions made by the decision fern and form the leaf nodes which are free to take any range of values. 

Hence, our final controller network is expressed as:
\begin{equation}
C(z_{t-n},...,z_{t-1},\theta) = z_{t-1} + TN(z_{t-n},...,z_{t-1},\theta)
\end{equation}

%===========================================================
\section{Experiments}
\label{sec:experiments}
In our experiments, we explore operation sets that can be achieved by working in the latent space. We choose two applications to focus on - the first application looks towards imposing a spatial transformation constraint on the latent vector; the second application is a more complex one, looking at video prediction. For the first application, we use the MNIST dataset~\cite{lecun1998mnist} as a toy example and investigate rotating and dilation operations on the dataset. For the second application, we use the KITTI dataset~\cite{geiger2013vision} and perform video prediction and steering prediction. For all experiments, we trained using the ADAM Optimiser~\cite{kingma2014adam} with learning rate of 0.0001, first and second moment values of 0.9 and 0.999 respectively, using a batch size of 64.

%------------------------------------------
\subsection{Imposing Spatial Transformation} 
\label{ssec:imposing_spatial}
For imposing spatial transformations, we present rotating and dilation operations and show how to constrain the direction a latent vector will traverse by using a spatial constraint. This constrained version of LSTNet applies a transformation to a single image which either rotates or erodes/dilates the given image. For both rotation and dilation experiments, we use a small residual VAE architecture along with our specified controller module with 1 residual layer with decision fern blocks (refer to Fig.~\ref{fig:controller_arch} \&~\ref{fig:fern_block} for details). We choose a latent vector size of 100 dimensions. The encoder consists of 2 residual downsampling layers (refer to~\cite{kingma2016improved} for details on the residual layers), Batch Normalisation and ReLU activations in between, ending with 2 fully connected linear layers for emitting the mean and variance of the latent vector. The decoder also consists of 2 residual upsampling layers, Batch Normalisation and ReLU activations in between, ending with a Hyperbolic Tangent function. To compare our method, we use two baselines. The first method is the most obvious comparison; we implement a baseline CNN which learns a target transformation, given an input image and corresponding control variable $\theta$ (CNN-baseline). This CNN-baseline is composed of 2 strided 3x3 convolution layers, 2 FC linear layers and 2 strided 3x3 deconvolution layers with ReLU non-linearities used for activation. The number of output channels in the hidden layers was kept at a constant 128. Additionally, we implement the Deep Convolutional Inverse Graphics Network~\cite{kulkarni2015deep} as specified in their paper for comparison. For each of the three methods compare under the same conditions by providing the control variable $\theta$ as an input during training and testing.
\paragraph{Rotation}
We create augmented, rotated samples from the MNIST dataset. Specifically, we randomly choose 600 samples from the data, ensuring a even distribution of 60 samples per class label are chosen. For each sample, we generate 45 rotation augmentations by rotating the sample in the range of $-45^{\circ}<\theta<45^{\circ}$. We add this augmented set to the original MNIST data and train the VAE with controller module end-to-end for 20k iterations. To inject the control variable into the input, we concatenate it to the encoded latent vector before feeding it to the controller module.

\begin{figure*}
\centering
 \subfloat[]{%
  \includegraphics[width=0.8\textwidth]{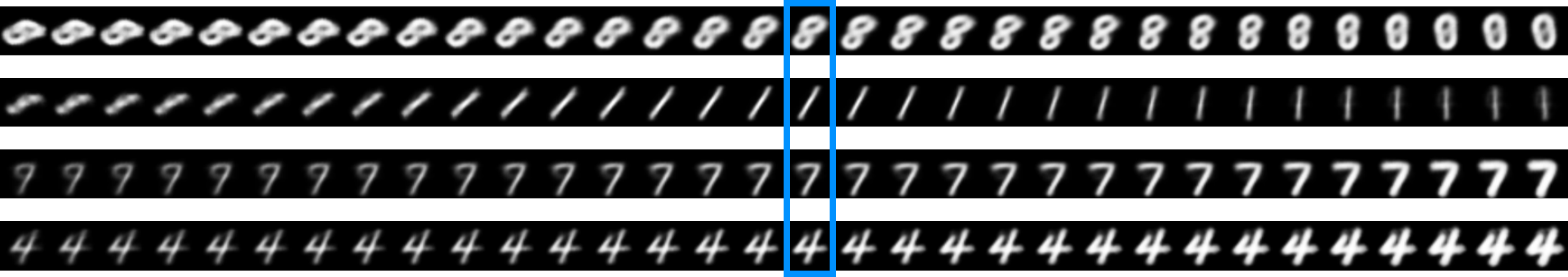}
  \label{fig:mnist_rot_dil_separate}
 }%
 \subfloat[]{%
  \includegraphics[width=0.14\textwidth]{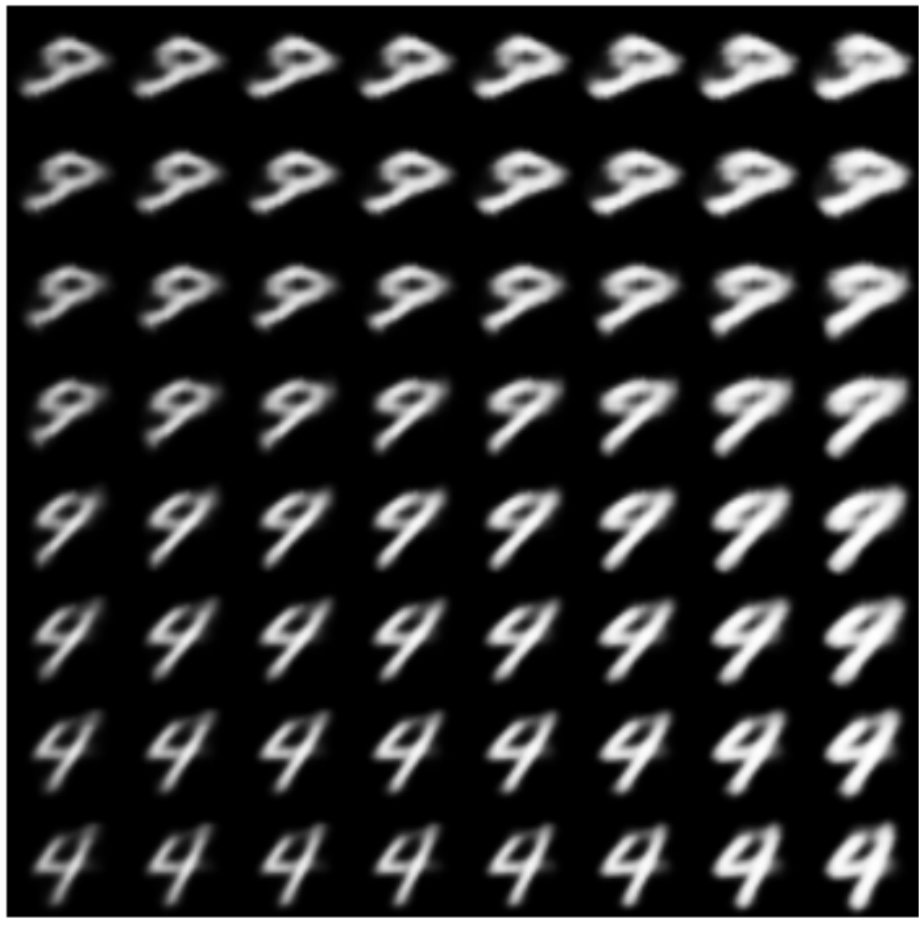}
  \label{fig:mnist_rot_dil_together}
 }%
\caption{(a) Rotation and dilation operations performed by LSTNet on samples from MNIST. The top two rows show rotation, whilst the bottom two rows show dilation. The original samples are shown in the middle highlighted in blue (b) LSTNet applying a combination of rotation and dilation operations to a sample from MNIST.}
\label{qual_results}
\end{figure*}

To train our controller module, we note that there are $\binom{45}{2}$ possible pairs in each example giving us much more training data than needed. Hence, for every iteration of training, we randomly choose a batch of triplets $(I_i,I_j,\theta)$, where $\theta$ is a rotation control variable specifying the rotation in radians. For inference, we randomly sample images from the MNIST dataset that were not selected to be augmented and perform a rotation parameter sweep. This results in smooth rotation of the image, while preserving the shape (see Fig.~\ref{fig:mnist_rot_dil_separate}). Note that other works (\textit{i.e}~\cite{chen2016infogan}) have shown that by altering a variable in the latent space, a rotated image can be retrieved, but fail to preserve image shape, leading to distortion. This indicates that the specific variable for rotation is not only responsible for rotation, but has influence over other attributes of the image. In contrast, we observed this difference across several variables between $z_i$ and $z_j$. This gives the insight that in order to perform a rotation (or similar spatial transform operation) and preserve the original image shape, several variables in the latent vector need to change which justifies the use of a highly non-linear network to approximate this operation.
%------------------------------------------
\paragraph{Thickening}
For learning the dilation operation we created augmentations in a similar manner to rotation operations. We randomly choose 5000 samples (evenly distributed across the 10 class labels) from the original dataset, and augmented every sample 4 times with 2 steps of dilation (thickening) and 2 steps of erosion (thinning). We train our VAE and controller module in a similar way to the rotation operation for 20k iterations, specifying batches of triplets $(I_i,I_j,\theta)$; $\theta$ changes its role to specifying a dilation factor that takes one of 5 discrete values in the range of $[-2,2]$. Note that although the network was trained on 5 discrete levels of dilation, it manages to learn to smoothly interpolate when performing the operation sweep during inference (as shown in Fig.~\ref{fig:mnist_rot_dil_separate}).
%------------------------------------------
\paragraph{Combining Operations}
One immediate extension that LSTNet offers when performing spatial transformation operations is the ease in which multiple spatial transformation operations can be combined together into a single framework. In Fig.~\ref{fig:mnist_rot_dil_together}, we show the samples produced by an LSTNet with 2 controller modules, sharing a single latent space offered by the VAE. 
% The whole framework is trained jointly end-to-end, this time with both rotation and dilation augmentation sets. 
It is important to note here that neither the rotation or dilation controller modules saw the other's training data. Hence, both operations are applied consecutively in the latent space and decoded back for visualisation. 

In Table~\ref{table:mnist_mse}, we show the mean squared error (MSE) of LSTNet, comparing against the two baseline methods across rotation, dilation and combined rotation plus dilation operations. We can see that LSTNet outperforms in Rotation and Dilation MSE and handily outperforms in the combined operation. This indicates a generality in learning in the latent space; LSTNet has learned the semantics behind rotation and dilation operations and thus can seamlessly combine these two operations.
\begin{table}[h]
\small
\begin{center}
\begin{tabular*}{\textwidth}{l@{\extracolsep{\fill}}ccc}
\hline
\thead{Model} &\thead{Rotation (MSE)} &\thead{Dilation (MSE)} &\thead{Rotation+Dilation (MSE)} \\
\hline
DCIGN~\cite{kulkarni2015deep} & 0.07373484 & 0.02599725 & 0.08349568 \\
CNN-baseline & 0.02819574 & 0.00841950 & 0.06508310 \\
LSTNet & \textbf{0.02380177} & \textbf{0.00835836} & \textbf{0.04410466} \\
\end{tabular*}
\caption{Mean squared error values on MNIST for rotation, dilation and rotation+dilation operations across CNN-baseline, DCIGN and LSTNet. Note that significant improvement LSTNet offers for the combined operation of rotation and dilation, indicating modularity.}
\label{table:mnist_mse}
\end{center}
\end{table} 
%------------------------------------------
\subsection{Imposing Kinematics} 
\label{ssec:imposing_kinematics}
We now move towards the more complex inference task of video prediction. Similar to imposing spatial transformation, we use a larger, residual VAE architecture along with a larger controller module to account for the more complex inference task. We increase the base dimension of our latent vector to 256 dimensions. The VAE's encoder consists of 3 residual downsampling layers, Batch Normalisation and ReLU activations in between, ending with a fully connected linear layer. The decoder consists of 3 residual upsampling layers, Batch Normalisation and ReLU activations in between, ending with a Hyperbolic Tangent function. For the controller network we test out two variants. The first is the fern-based controller as described in Section~\ref{ssec:lstnet}. For motivating the use of the fern-based controller, a linear variant controller network is also used: this is a feedforward linear network consisting of 4 FC linear layers with ReLU activations, matching the fern-based controller in terms of model capacity.

Similarly to imposing spatial transformations, we randomly create batches of triplets $(I_i,I_j,\theta)$, where $I_i$ and $I_j$ are the respective current frame and target future frame (randomly chosen to be within 5 time steps of the current frame) and $\theta$ is the corresponding time step from the current frame to the target frame. Our controller module receives the latent vectors of the current frame as well as a sequence of latent vectors belonging to the previous 5 frames to the current frame. We train our framework end-to-end for 150k iterations; the VAE is trained to minimise reconstruction loss along with KL regularisation, the controller is minimised using the latent error between the target latent and predicted latent vectors and both decoder and controller are jointly optimised using the error between target frames and predicted frames (refer to Eqs.~\ref{eq:vae_loss},~\ref{eq:controller_loss} and~\ref{eq:prediction_loss}). The VAE is trained at a ratio of 5:1 against the controller module to ensure a proper representation of the input in its encoding space.

\begin{figure*}
	\begin{center}
		\includegraphics[width=0.95\linewidth]{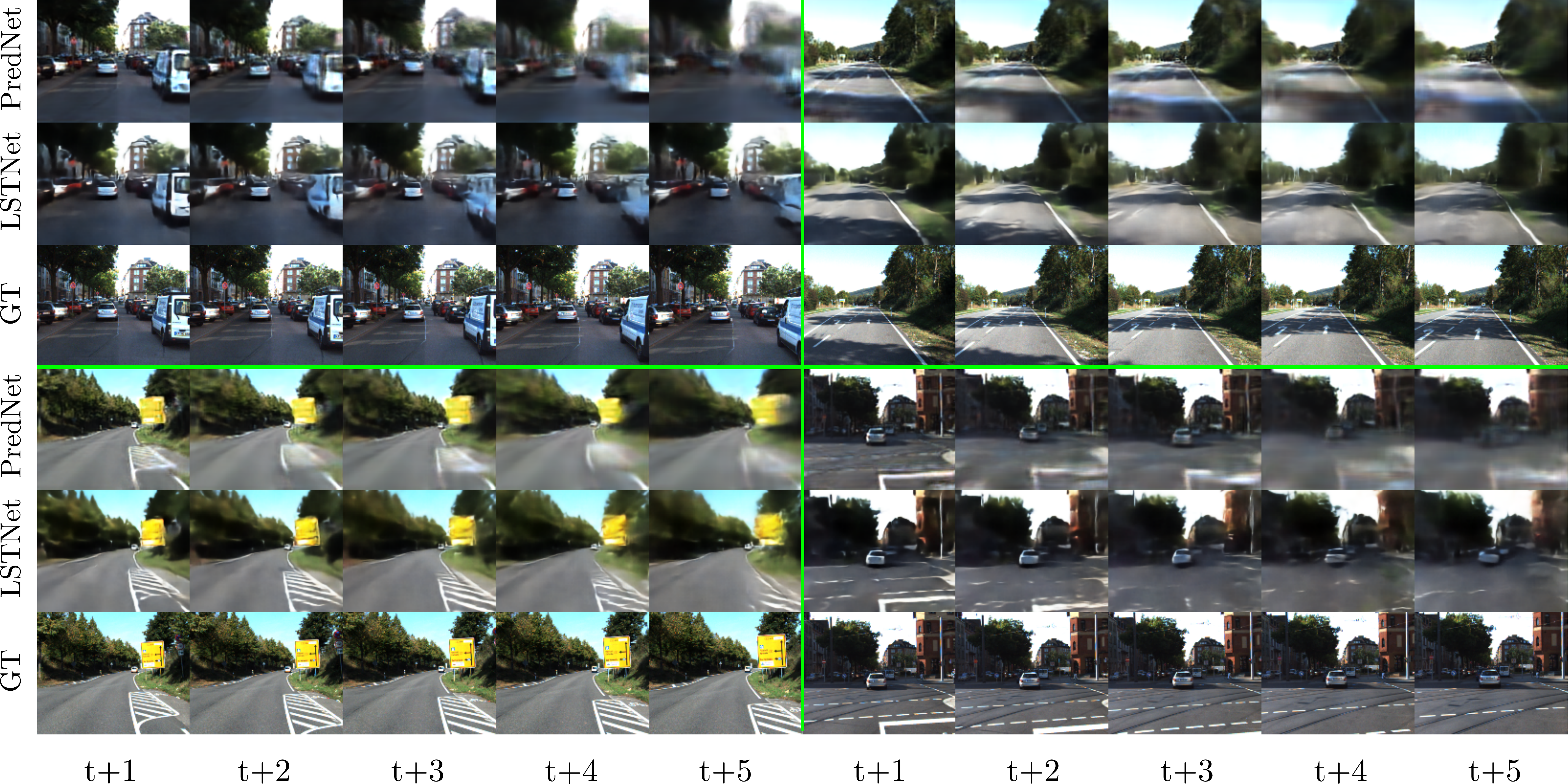}
	\end{center}
	\caption{Multi-Frame predictions. This figure depicts of 4 sequences selected from the KITTI test set where $\{I_{t+1},I_{t+2},I_{t+3},I_{t+4},I_{t+5}\}$ are predicted using a past sequence of 5 frames.}
\label{fig:qual_results}
\end{figure*}

Fig.~\ref{fig:qual_results} shows qualitative results comparing our model with the PredNet model in~\cite{lotter2016deep}. We can see that the further the prediction is over time, the less accurate PredNet becomes, whilst LSTNet remains much more robust to changes in the scene over time. Observing the samples of PredNet, a recurring phenomena is that in areas of the frame where object movement should occur, moving objects are instead smeared and blurred. In the case of LSTNet, predicted movement is better observed (particularly over large time steps), whilst vehicles and street furniture (\textit{i.e} road signs and markings) are better placed. This gives an indication that LSTNet is more reliable for prediction mechanism within the context of the task. PredNet implicitly learns to predict on a fixed timestep and hence predicts over a large time interval by rolling out over its predicted images. On the other hand, LSTNet offers a more general approach via the control variable $\theta$. It learns a transformation which allows it to directly predict the target video frame without the computational overhead of rolling out fixed timesteps to reach the target.
\begin{figure}
\centering
 \hskip -1ex
 \subfloat[]{%
 \begin{tabular}{l c c}
	\hline
	\thead{Model} &\thead{Avg. MSE} &\thead{Avg. SSIM} \\
	\hline
  	Copy Last Frame & 0.03829 & 0.615 \\
  	PredNet [7] & 0.02436 & 0.604 \\
  	PredNet (Finetuned) [7] & 0.01524 & 0.679 \\
  	Linear Controller Variant & 0.02083 & 0.631 \\
  	LSTNet (Ours) & \textbf{0.01316} & \textbf{0.694} 
  \end{tabular}
  \label{fig:kitti_mse_table}
 }%
 \subfloat[]{%
  \begin{tabular}{c}
  \includegraphics[width=0.4\textwidth]{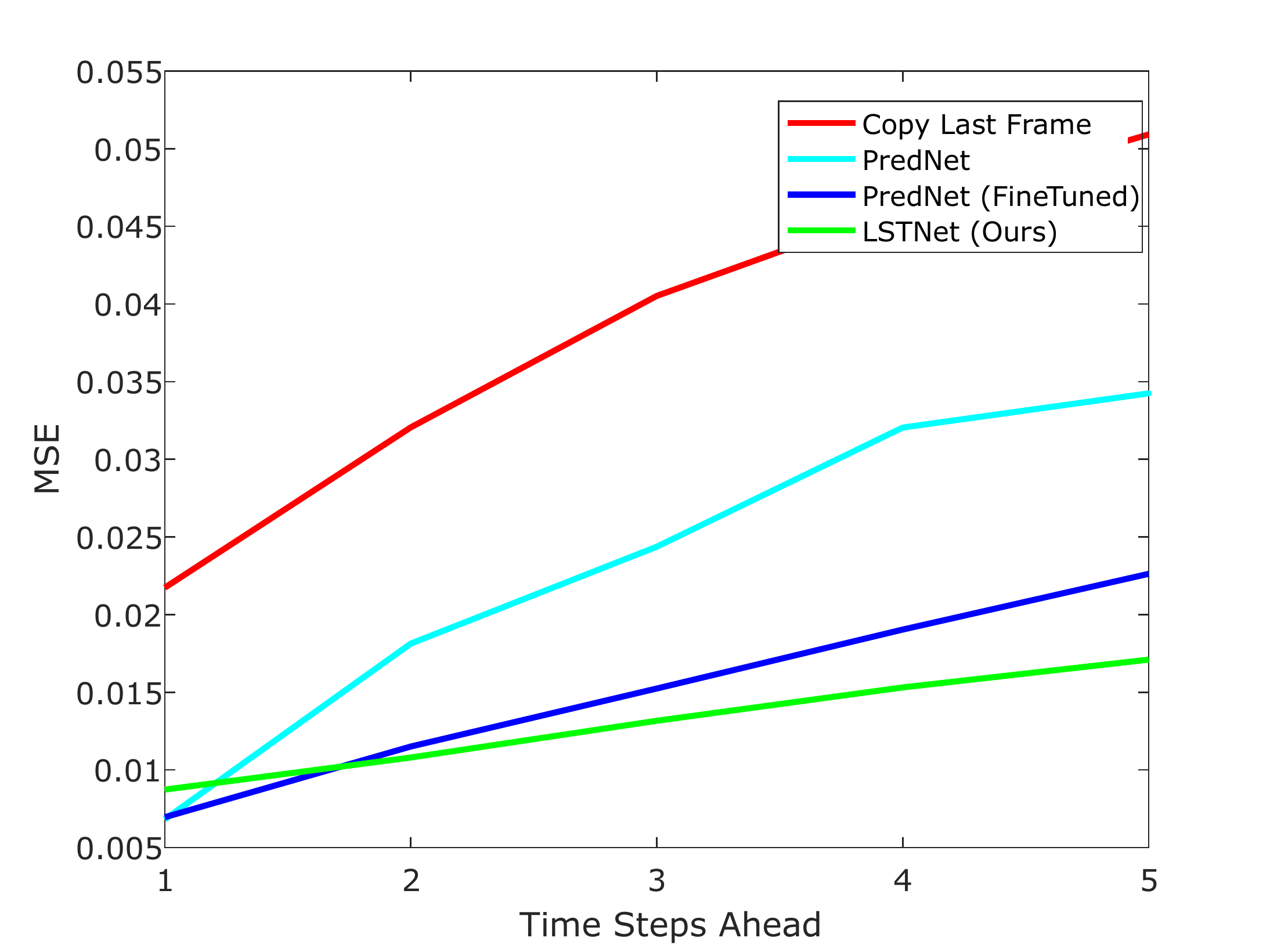}
  \end{tabular}
  \label{fig:kitti_mse_plot}
 }%
\caption{(a) Averaged MSE and SSIM over 5 timesteps of frame prediction (b) Individual MSE for each predicted time step plotted for Copy Last Frame, PredNet and LSTNet.}
\end{figure}
In Table~\ref{fig:kitti_mse_table}, we depict the average MSE and SSIM over 5 future frames (500ms time lapse) for the KITTI test set. We can see that LSTNet outperforms the compared methods of Copy Last Frame, PredNet and a Linear Controller Variant (as discussed in Section~\ref{ssec:lstnet}). Looking at Fig.~\ref{fig:kitti_mse_plot}, we can see that LSTNet achieves lowest MSE for all timesteps, except $\hat{I}_{t+1}$ , where PredNet has slightly lower MSE. However, despite slightly higher initial MSE, the performance of LSTNet quickly exceeds both methods as inferring good prediction begins playing a larger factor over time (see Section~\ref{ssec:latent_space_for_prediction} for a full discussion). These quantitative results correlate well with our qualitative results and again indicate that LSTNet is able to outperform on the task of prediction rather than on image reconstruction.

\subsection{Latent Space for Prediction} 
\label{ssec:latent_space_for_prediction}
In addition to the computational and memory footprint benefits, we show that projecting an image onto the latent space, operating on the latent vector using our controller module and reprojecting back into the pixel space has on average a lower MSE in the pixel space over operations that are increasingly harder to perform; for example: chaining spatial transformations and predicting more than 1 time step into the future. 

\begin{figure*}[h]
	\begin{center}
		\includegraphics[width=0.9\textwidth]{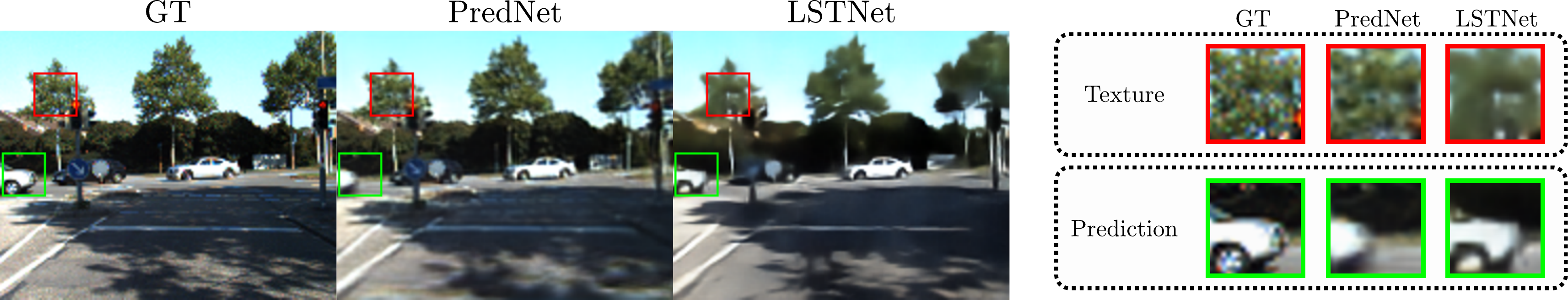}
	\end{center}
	\caption{On the left hand side we present the next frame prediction ($t+1$) for the Ground-Truth, PredNet (Finetuned) and our LSTNet. On the right hand side are patches that match the rectangular markings on the images with corresponding labels. Our LSTNet excels at predicting the semantic changes that are important for maintaining the correct structure of a scene; and at times may fail (as shown) at outputting the fine-grained details of the scene objects, due to reconstruction.}
\label{fig:patch_compare}
\end{figure*}

\paragraph{Moving vs. non-moving objects}
An attribute that does not favour a latent space framework is the inference of the fine grained detail in images with a lot of static, non-moving components. Fig.~\ref{fig:patch_compare} shows an example of a $\hat{I}_{t+1}$ prediction for PredNet (Finetuned)~\cite{lotter2016deep}, our LSTNet and the ground truth for comparison. Across these three images, red and green highlighted patches show texture and movements of objects respectively. Across the patches, it is visually apparent that LSTNet does not generate a fine detailed texture of the tree leaves, although it captures the movement of a car well as the camera viewpoint changes. Conversely, PredNet is able to capture the fine grain texture of the tree leaves, but fails to capture the movement of the stationary car from a camera viewpoint change.

We perform a simple patch-based test to illustrate the importance of emphasising prediction on moving objects versus non-moving objects. For $\hat{I}_{t+1}$, we randomly choose 20 patches of 20$\times$20 pixels that we identify where movement occurs; computing the average MSE yielded values for LSTNet=\textbf{0.0020} and PredNet=0.0134 showing we significantly improve on predicting movement. Similarly, we sample 20 patches identified as being static with no moving objects which yielded an MSE for LSTNet=0.0024 and PredNet=\textbf{0.0021}, with the results favouring PredNet. This result correlates well with the competitive results shown on $\hat{I}_{t+1}$ in Fig.~\ref{fig:kitti_mse_plot}; between two consecutive frames, the majority of parts in a scene are static with little to no movement. Hence, prediction plays a smaller role in pixel space MSE over such a short time frame. However, as the time between predicted and current frame increases, getting better predictions on movements plays a larger role, which accounts for the results shown in Fig.~\ref{fig:kitti_mse_plot}.
\paragraph{Auxiliary Parameter Predictions}
Furthermore, inferring auxiliary tasks such as steering angle does not require the fine detailed knowledge contained in the pixel space of a scene. For performing such inference tasks, the main requirement is the semantic information contained in the scene. For this task, we used a pretrained LSTNet and added a FC layer to the controller to output a single value and finetuned to learn the steering angle. This is where LSTNet shines; it considerably outperforms PredNet~\cite{lotter2016deep} and copying steering angles from the last seen frame as shown in Table~\ref{fig:kitti_steering_mse}. This further correlates with our MSE prediction results that LSTNet is inherently able to distill the semantics from a scene for complex inference tasks.   
\begin{table}
\centering
  \begin{tabular}{l c}
  \hline
  \thead{Model} &\thead{Steering Angle MSE \\ (Degrees$^2$)} \\
  \hline
  Copy Last Frame & 1.3723866  \\
  PredNet (FineTuned)~\cite{lotter2016deep} & 2.4465750 \\
  LSTNet (Ours) & \textbf{0.5267124} \\
  \end{tabular}
\caption{Steering angle prediction MSE on the KITTI test data.}
\label{fig:kitti_steering_mse}
\end{table}

\section{Conclusion}
In this work, we present a novel, end-to-end trainable framework for operating on a latent space constructed using a VAE. We explore the power of learning in latent space on two operations: spatial transformations and video prediction, and show semantic gains for increasingly harder inference tasks which subsequently translates to a more meaningful result in the pixel space. Furthermore, as a direct extension to this work, the use of a VAE presents an opportunity to explore multi-model predictions for further robustness in predictive tasks.

%===========================================================

\end{document}